\documentclass[sigconf]{acmart}

\copyrightyear{2025}
\acmYear{2025}
\setcopyright{cc}
\setcctype{by}
\acmConference[SIGIR '25]{Proceedings of the 48th International ACM SIGIR Conference on Research and Development in Information Retrieval}{July 13--18, 2025}{Padua, Italy}
\acmBooktitle{Proceedings of the 48th International ACM SIGIR Conference on Research and Development in Information Retrieval (SIGIR '25), July 13--18, 2025, Padua, Italy}
\acmDOI{10.1145/3726302.3730299}
\acmISBN{979-8-4007-1592-1/2025/07}

\usepackage{multirow}
\usepackage{pifont}
\usepackage{xspace}
\usepackage{enumitem}
\usepackage{tabularx}

\definecolor{customblue}{rgb}{0.168, 0.364, 0.557}
\colorlet{framegray}{gray!3!white}
\newcommand{\datasetname}{\textsc{PlausibleQA}\xspace}

\newcommand{\blackcircle}[1]{%
    \raisebox{0.8pt}{\textcircled{\scriptsize\raisebox{-0pt}{#1}}}
}

\begin{document}

\title{Wrong Answers Can Also Be Useful: \datasetname — A Large-Scale QA Dataset with Answer Plausibility Scores
}

\author{Jamshid Mozafari}
\orcid{0000-0003-4850-9239}
\affiliation{%
	\institution{University of Innsbruck}
	\city{Innsbruck}
	\state{Tyrol}
	\country{Austria}
}
\email{jamshid.mozafari@uibk.ac.at}

\author{Abdelrahman Abdallah}
\orcid{0000-0001-8747-4927}
\affiliation{%
	\institution{University of Innsbruck}
	\city{Innsbruck}
	\state{Tyrol}
	\country{Austria}
}
\email{abdelrahman.abdallah@uibk.ac.at}

\author{Bhawna Piryani}
\orcid{0009-0005-3578-2393}
\affiliation{%
	\institution{University of Innsbruck}
	\city{Innsbruck}
	\state{Tyrol}
	\country{Austria}
}
\email{bhawna.piryani@uibk.ac.at}

\author{Adam Jatowt}
\orcid{0000-0001-7235-0665}
\affiliation{%
	\institution{University of Innsbruck}
	\city{Innsbruck}
	\state{Tyrol}
	\country{Austria}
}
\email{adam.jatowt@uibk.ac.at}



\begin{abstract}
	Large Language Models (LLMs) are revolutionizing information retrieval, with chatbots becoming an important source for answering user queries. As by their design, LLMs prioritize generating correct answers, the value of highly plausible yet incorrect answers (candidate answers) tends to be overlooked. However, such answers can still prove useful, for example, they can play a crucial role in tasks like Multiple-Choice Question Answering (MCQA) and QA Robustness Assessment (QARA). Existing QA datasets primarily focus on correct answers without explicit consideration of the plausibility of other candidate answers, limiting opportunity for more nuanced evaluations of models. To address this gap, we introduce \datasetname, a large-scale dataset comprising 10,000 questions and 100,000 candidate answers, each annotated with plausibility scores and justifications for their selection. Additionally, the dataset includes 900,000 justifications for pairwise comparisons between candidate answers, further refining plausibility assessments. We evaluate \datasetname through human assessments and empirical experiments, demonstrating its utility in MCQA and QARA analysis. Our findings show that plausibility-aware approaches are effective for MCQA distractor generation and QARA. We release \datasetname as a resource for advancing QA research and enhancing LLM performance in distinguishing plausible distractors from correct answers.
\end{abstract}




\maketitle

\section{Introduction}\label{s:introduction}

\begin{figure*}
	\centering
	\includegraphics[width=0.75\textwidth]{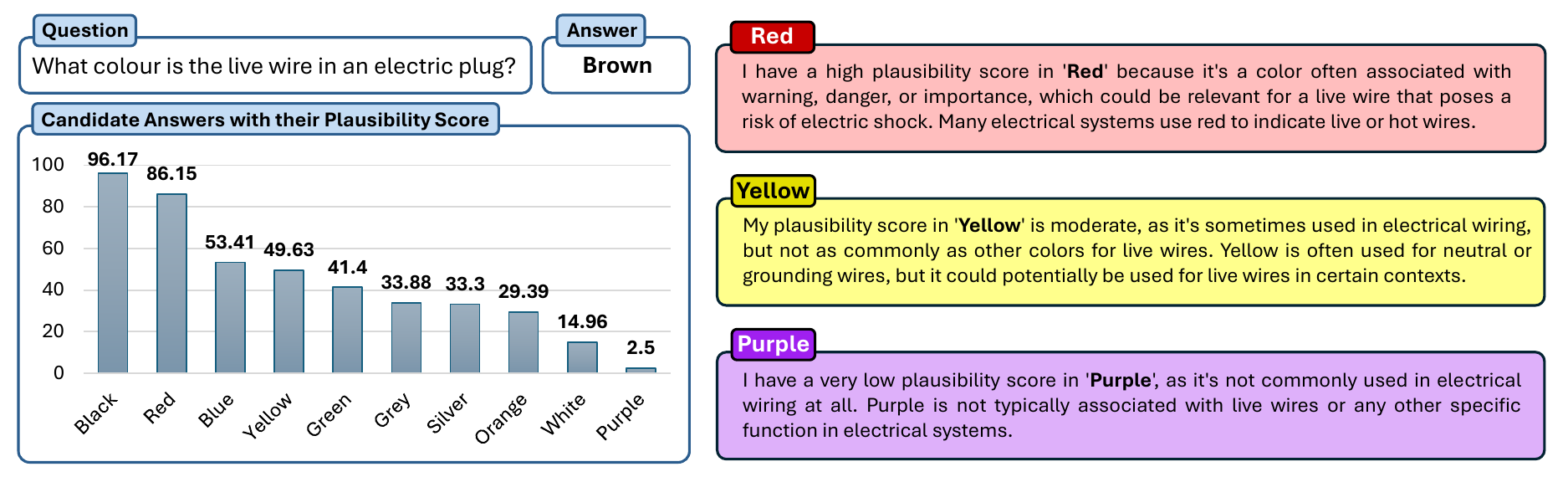}
	\caption{Example from the \datasetname dataset: The \texttt{Question} and \texttt{Answer} sections present the question along with its gold answer. The \texttt{Plausibility Score} section ranks candidate answers in descending order based on their plausibility scores. \texttt{Justification Boxes} provide the justifications for the selection of each candidate answer.}
	\label{fig:sample}
        \Description{}
\end{figure*}

Question Answering (QA) systems have become increasingly important in recent years, enabling users to efficiently obtain answers to a wide range of questions~\cite{10.1007/s10115-022-01783-5, mozafari-etal-2024-exploring, INR-102, zhao-etal-2021-sparta}. The rapid advancement of Large Language Models (LLMs)~\cite{2023arXiv230308774O, 2024arXiv240305530G, 2024arXiv241219437D} has significantly contributed to this progress, impacting various natural language processing (NLP) tasks~\cite{2024arXiv240512819Q}. Most users expect QA systems to provide correct answers, and LLMs have made accessing accurate information more convenient~\cite{kamalloo-etal-2023-evaluating}. However, the role of incorrect yet plausible answers is often overlooked, despite their potential usefulness in various applications~\cite{pelanek2016properties}. For instance, in multiple-choice question answering (MCQA) and educational assessment, plausible distractors are crucial for constructing meaningful answer options~\cite{2025arXiv250113125L}. Similarly, contrastive learning tasks and robustness assessment of QA models (QARA) can benefit from high-quality incorrect answers~\cite{cao-etal-2022-exploring}.

A key approach to generating high-quality incorrect answers is through the creation of \textit{candidate answers}\footnote{In this study, we use the term \textit{candidate answers} to refer to \textit{plausible yet incorrect answers} for questions. This means that the correct answer is not included.}
. Candidate answers can be generated using retrieval-based methods, which extract answers from retrieved passages or sentences~\cite{yu-etal-2024-enhancing}, generative approaches that leverage LLMs to produce plausible answers~\cite{qu-etal-2024-unsupervised}, and human-curated methods where annotators generate plausible answers~\cite{lai-etal-2017-race}. However, merely generating candidate answers is insufficient; their quality plays a crucial role in downstream applications~\cite{alhazmi-etal-2024-distractor}. A major limitation of existing methods is their binary treatment of answers—only classifying them as either correct or incorrect without considering plausibility~\cite{raina-etal-2023-assessing}. A \textit{plausibility score}, on the other hand, would quantify how likely a candidate answer appears to be correct, independent of its actual correctness. For example, in response to the question \textit{What is the capital of the USA?}, \textit{New York} and \textit{Memphis} are incorrect, but \textit{New York} is more plausible as a potential capital than \textit{Memphis} due to its higher prominence and so on. In this case, the system could assign \textit{New York} a high plausibility score, for instance, $0.95$, while it could give to \textit{Memphis} a low score such as $0.15$. In practice, for example, in QARA~\cite{han-etal-2023-robustqa}, high-plausibility candidate answers provide a more challenging and informative robustness evaluation compared to using obviously incorrect answers. Similarly, in MCQA systems, answer plausibility can be used to calibrate question difficulty, selecting highly plausible candidate answers for questions deemed to be difficult and less plausible ones for questions that should be easy.

In this paper, we present \datasetname\footnote{\url{https://github.com/DataScienceUIBK/PlausibleQA}}, the first large-scale dataset explicitly featuring candidate answers with plausibility scores. The dataset comprises 10,000 questions and 100,000 candidate answers, each annotated with a plausibility score reflecting its perceived likelihood of correctness. Additionally, we provide detailed justifications for the selection of each candidate answer and its assigned plausibility score. Furthermore, the dataset includes 900,000 justifications for pairwise comparisons between candidate answers within each question. Figure~\ref{fig:sample} provides an example showcasing a question, its correct answer, ten candidate answers, their associated plausibility scores, and justifications for three candidate answers. In the experiments section, we utilize \datasetname to examine the role of plausibility scores in MCQA distractor generation and to analyze QA robustness of LLMs.

In summary, we make the following contributions in this study:
\begin{itemize}  
	\item We introduce \datasetname, the first large-scale dataset incorporating plausibility scores for candidate answers, consisting of 10,000 questions, 100,000 annotated candidate answers, and 1,000,000 justifications.
	\item We validate the \datasetname dataset by conducting a human evaluation on the dataset.
	\item We perform comprehensive analyses and experiments, which attest to high quality of \datasetname and we demonstrate its applicability across multiple QA-related tasks, including MCQA and QARA. 
\end{itemize}

The remainder of this paper is organized as follows: Section~\ref{s:related_work} reviews existing work on QA, MCQA, and QARA, highlighting limitations and the necessity of plausibility scores. Section~\ref{s:dataset} describes the dataset construction process and key attributes.  Section~\ref{s:dataset_analysis} provides a comprehensive analysis of the dataset, including statistical analysis and human evaluation results. Section~\ref{s:experiments} presents experimental findings on MCQA and QARA. Finally, Section~\ref{s:conclusion} discusses the broader impact of \datasetname and suggests directions for future research.

\begin{figure*}
	\centering
	\includegraphics[width=0.75\textwidth]{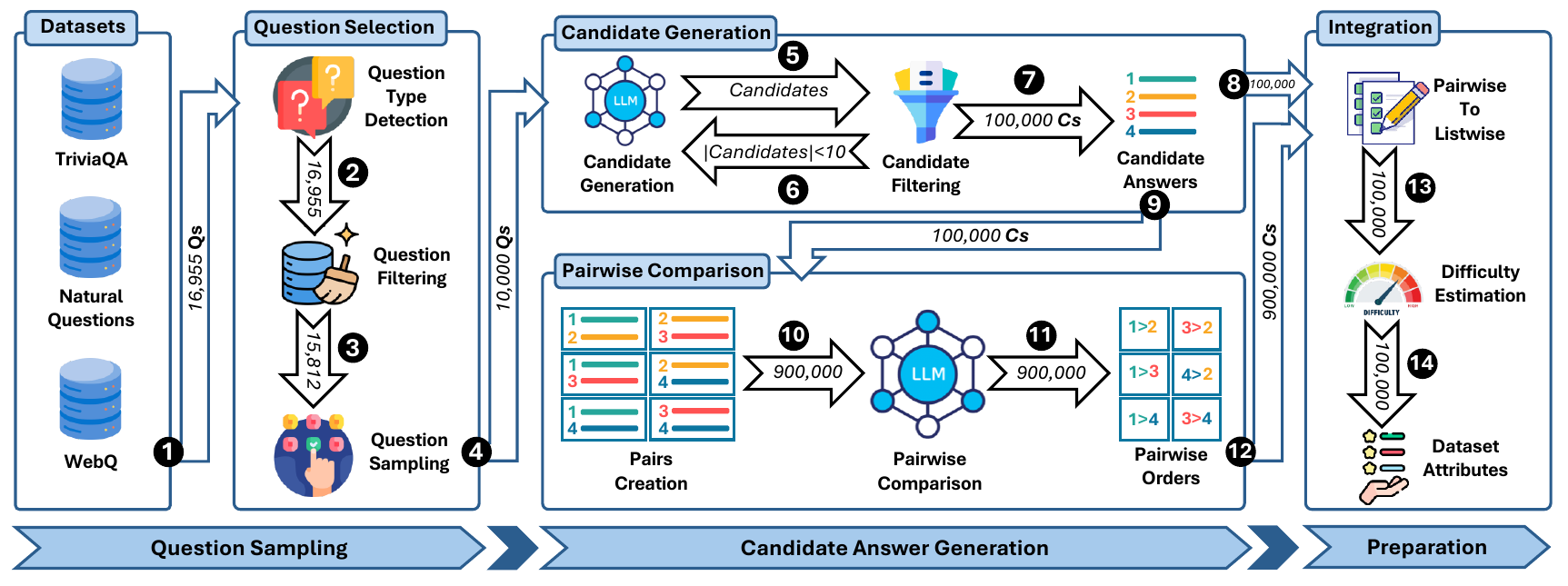}
	\caption{The pipeline of \datasetname generation: \blackcircle{1}Datasets are passed to the Question Sampling. \blackcircle{2}The type of each question is detected. \blackcircle{3}Questions are filtered accordingly. \blackcircle{4}10,000 questions are sampled from the filtered questions. \blackcircle{5}Using LLMs, 10 candidate answers are generated for each question. \blackcircle{6}If the output does not pass the filtering stage, it is retried until \blackcircle{7}10 unique candidate answers are generated. \blackcircle{8}The candidate answers are then passed to the Preparation stage. \blackcircle{9}Before this, the candidate answers undergo Pairs Creation.
		\blackcircle{10}The candidate answers are converted into pairwise items \blackcircle{11}and passed to an LLM for pairwise comparison . \blackcircle{12}The results are passed to the Preparation stage, where they are converted from pairwise order to listwise order. \blackcircle{13}The question difficulty and answer difficulty are then evaluated. Finally, \blackcircle{14}all generated attributes are stored in the \datasetname dataset. The numbers in the arrows indicate the number of output items. Additionally, \texttt{Qs} represents \textit{Questions}, and \texttt{Cs} represents \textit{Candidates}.}
	    
	\label{fig:pipeline}
        \Description{}
\end{figure*}

\section{Related Work}\label{s:related_work}
In this section, we review prior work on QA, MCQA, and QARA. While substantial progress has been made in these areas, most existing approaches lack explicit mechanisms for leveraging and evaluating the plausibility of candidate answers.

\subsection{QA Task}
QA has long been a core task in IR/NLP, leading to the development of numerous datasets designed to evaluate and enhance QA models~\cite{serban-etal-2016-generating, 2017arXiv170405179D, reddy-etal-2019-coqa, kwiatkowski-etal-2019-natural}. Early datasets, such as SQuAD~\cite{rajpurkar-etal-2016-squad}, focus on extractive QA, where models locate text spans within a passage to form answers. More recent datasets, including CoQA~\cite{reddy-etal-2019-coqa} and HotpotQA~\cite{yang-etal-2018-hotpotqa}, expand upon this by incorporating conversational QA and multi-hop reasoning. While these datasets prioritize factual correctness, they largely overlook the role of candidate answers in model evaluation.
Generative models have further transformed QA by producing free-form answers rather than relying solely on retrieval-based or extractive approaches~\cite{lewis-etal-2020-bart, NEURIPS2020_1457c0d6}. LLMs such as GPT-4~\cite{2023arXiv230308774O}, T0~\cite{sanh2022multitask}, and T5~\cite{10.5555/3455716.3455856} generate fluent and contextually appropriate responses but do not account for candidate answers and focus only on correct answers.

To address this gap, our proposed pipeline extends existing QA datasets by generating candidate answers for questions and incorporating plausibility scores for candidate answers. This approach enables a more nuanced evaluation of QA models, moving beyond a binary correctness framework and offering deeper insights into model reasoning.

\subsection{QA Robustness Assessment}
QA robustness refers to a model’s ability to maintain accuracy across different types of adversarial inputs or variations in question phrasing~\cite{jia-liang-2017-adversarial,Kaushik2020Learning}. Prior work has focused on adversarial QA datasets such as AddSent~\cite{jia-liang-2017-adversarial}, which introduces misleading sentences into contexts to challenge models. Similarly, ContrastiveQA~\cite{gardner-etal-2020-evaluating} evaluates robustness by modifying questions while keeping the correct answers unchanged. 

However, existing robustness studies often rely on explicit adversarial attacks such as manipulating input data to deliberately mislead models, altering text features to create confounding effects, or introducing subtly modified but incorrect information rather than naturally plausible wrong answers, which can also serve as a robustness test. \datasetname provides candidate answers with plausibility scores, allowing a more fine-grained robustness evaluation by measuring how well models distinguish between the correct answer and highly plausible candidate answers.

\subsection{MCQA Task}
MCQA requires models to select the correct answer from a set of options, often including distractors designed to challenge the model~\cite{lai-etal-2017-race,2018arXiv180305457C}. Datasets like RACE~\cite{lai-etal-2017-race} and ARC~\cite{2018arXiv180305457C} contain human-crafted distractors, while others, such as SciQ~\cite{welbl-etal-2017-crowdsourcing}, are generated automatically from scientific texts, with distractors derived from retrieved passages to improve model comprehension.

Despite their effectiveness, most MCQA datasets do not rank distractors based on plausibility. This limits the ability to generate adaptive MCQA tasks based on difficulty levels. In contrast, our dataset incorporates plausibility scores, enabling dynamic MCQA assessments that adjust distractor selection based 
on difficulty.

\section{\datasetname Dataset}\label{s:dataset}
This section presents the pipeline and its stages for generating the \datasetname dataset. The pipeline consists of three key modules: \textit{Question Sampling}, \textit{Candidate Answer Generation}, and \textit{Preparation}. Figure~\ref{fig:pipeline} visually illustrates the pipeline and its stages, which we will explore in detail in the following sections.

\subsection{Question Selection}\label{ss:question_selection}
In this study, we use three popular QA datasets—TriviaQA \cite{joshi-etal-2017-triviaqa}, Natural Questions (NQ)~\cite{kwiatkowski-etal-2019-natural}, and WebQuestions (WebQ)~\cite{berant-etal-2013-semantic}—as the source of questions. These datasets are extensively used as benchmarks in QA research. We focus only on the test sets of these datasets, which contain a total of 16,995 questions.

\subsubsection{Question Type Detection}\label{sss:question_type_detection}
A question classifier categorizes natural language questions based on their intended purpose. For example, \textit{Who is the president of Austria?} falls under the \textit{PERSON} class, while \textit{What is the capital of Austria?} belongs to the \textit{LOCATION} class. We fine-tune the RoBERTa~\cite{2019arXiv190711692L} model on the TREC Question Classification dataset~\cite{li-roth-2002-learning} to develop a model called \textit{QT Classifier} as a question classifier\footnote{The accuracy of QT Classifier on the TREC Question Classification dataset is $92.8\%$.}. The QT Classifier classifies questions at different levels of granularity, encompassing 6 coarse-grained classes and 50 fine-grained classes.

\subsubsection{Question Filtering}\label{sss:question_filtering}
The coarse-grained classes consist of HUM (Human), LOC (Location), ENTY (Entity), NUM (Number), ABBR (Abbreviation), and DESC (Description). We exclude questions classified under the DESC type, focusing only on factoid questions. Additionally, due to the limited number of instances, we merge NUM and ABBR into a single category labeled OTHER. As a result, we obtain a final set of 15,812 questions.

\subsubsection{Question Sampling}\label{sss:question_sampling}
We employ the stratified sampling method \cite{ARNAB2017213} to sample questions based on their coarse-grained classes. Questions are first grouped according to their type, and random sampling is then performed within each subgroup. In total, 10,000 questions (two-thirds) are selected. Finally, we shuffle them to ensure randomness.

\begin{figure}[t]
	\centering
	
	\begin{center}
		\fcolorbox{customblue}{framegray}{
			\begin{minipage}{0.88\linewidth}
				\scriptsize
				Assume that you are unaware that the answer to $<$question$>$ is $<$ground\_truth$>$. Generate a list of 10 unique candidate answers, ensuring that $<$ground\_truth$>$ is excluded. A plausibility score evaluates how reasonable, credible, or contextually appropriate each candidate answer is in relation to the given question. For each candidate, provide:
				\begin{enumerate}
					\item A non-zero plausibility score as a number between 0 and 100.
					\item A detailed explanation of the reasoning behind the plausibility score.
				\end{enumerate}
				
				Format your response as a JSON list, where each candidate is represented as:
				
				\begin{flushleft}
					\ttfamily
					[\\
						\ \ \{\\
						\ \ \ \ \ "CandidateAnswer": "<candidate\_answer>",\\
						\ \ \ \ \ "PlausibilityScore": <plausibility\_score>,\\
						\ \ \ \ \ "Justification": "<justification>"\\
						\ \ \}\\
					]
				\end{flushleft}
				\vspace{\baselineskip}
				The output must be a valid JSON list only.
			\end{minipage}
		}
	\end{center}
	\caption{Prompt for Candidate Answer Generation. \texttt{<question>} represents the given question, while \texttt{<ground\_truth>} denotes its correct answer. Each candidate answer is represented by \texttt{<candidate\_answer>}, with an initial plausibility score (\texttt{<plausibility\_score>}) and a justification (\texttt{<justification>}) justifying both the answer choice and the assigned plausibility score.}
	\label{fig:listwise-prompt}
	\Description{}
\end{figure}

\subsection{Candidate Answer Generation}\label{ss:candidate_answer_generation}
To generate candidate answers and compare pairs, we use LLaMA-3.3-70B~\cite{2024arXiv240721783G} as the LLM, as it is a highly capable model with strong performance in various NLP tasks. Also, it is open-source and available on HuggingFace\footnote{\url{https://huggingface.co/meta-llama/Llama-3.3-70B-Instruct}}, enabling users to reimplement our proposed pipeline and generate candidate answers for their custom datasets.

This module first generates candidate answers along with their listwise-based plausibility scores in the \textit{Candidate Generation} stage and then performs pairwise comparisons of the generated candidate answers in the \textit{Pairwise Comparison} stage, which we describe in the following sections. The results from the \textit{Pairwise Comparison} stage, combined with the listwise-based plausibility scores from the \textit{Candidate Generation} stage, are used to refine and produce more accurate plausibility scores, which we explain in Section~\ref{ss:preparation}.

\subsubsection{Candidate Generation}\label{sss:candidate_generation}

We prompt the LLM to generate 10 unique candidate answers\footnote{We tested various values and found 10 to be optimal.} for each question, assigning listwise-based plausibility scores between 0 and 100. To ensure unbiased scoring, the model is explicitly instructed to exclude the ground truth answer, as including it skews the listwise-based plausibility score distribution by assigning a score close to 100 to the correct answer while giving significantly lower scores to candidate answers. Additionally, we require the LLM to explain why each candidate answer was selected and how its listwise-based plausibility score was determined, which helps improve the quality of generated candidate answers and plausibility score assessments~\cite{huang-etal-2023-large}. Finally, the results are structured as a list of JSON objects for consistency and ease of processing. The prompt is illustrated in Figure~\ref{fig:listwise-prompt}.

\begin{figure}[t]
	\centering
	
	\begin{center}
		\fcolorbox{customblue}{framegray}{
			\begin{minipage}{0.88\linewidth}
				\scriptsize
				Evaluate the two candidate answers below and determine which one is more likely to be correct based on the given question. A 'better candidate answer' is the one with a higher probability of being correct. Assume that you do not know the correct answer is $<$ground\_truth$>$. After evaluating, provide your response.
				\newline
						
				Question: $<$question$>$
						    
				Candidate Answer 1: $<$ca\_1$>$
						    
				Candidate Answer 2: $<$ca\_2$>$
				\newline
						
				Which candidate answer is better?
			\end{minipage}
		}
	\end{center}
	    
	\caption{Prompt for Pairwise Candidate Answer Comparison. \texttt{<question>} represents the given question, while \texttt{<ground\_truth>} denotes its correct answer. The candidate answers $x_1$ and $x_2$ are represented by \texttt{<ca\_1>} and \texttt{<ca\_2>}, respectively.}
	\label{fig:pairwise-prompt}
	\Description{}
\end{figure}

Next, the generated JSON list is passed to the \textit{Candidate Filtering} stage, where we validate the candidate answers by: (1) ensuring no duplicate candidate answers are present, (2) verifying that listwise-based plausibility scores fall within the valid range, and (3) confirming that exactly 10 unique candidate answers are generated. To detect duplicate answers, we employ the BERT Matching (BEM) method~\cite{bulian-etal-2022-tomayto} for semantic similarity comparison\footnote{We treat candidate answers with more than 85\% similarity as identical.}. If any conflicts arise, we increase the LLM’s temperature by 0.1 and prompt the model to regenerate. This process is repeated iteratively until the generated list successfully passes the filtering stage. Once all questions have been processed through the filtering stage, we obtain a total of 100,000 candidate answers for 10,000 questions.

\subsubsection{Pairwise Comparison}\label{sss:pairwise_comparison}
To assess the reliability and accuracy of the listwise-based plausibility scores generated during the \textit{Condidate Generation} stage, we compute plausibility scores through pairwise comparisons of candidate answers, referred to as pairwise-based plausibility scores. We then analyze the correlation between the listwise-based and pairwise-based plausibility scores to determine their alignment.

To this end, we generate pairwise comparisons for the candidate answers while ensuring that no item is paired with itself. Formally, given $\mathcal{C}_q$ as the list of candidate answers for a question $q$, we construct the set of pairs as $\{(x_1, x_2) \mid x_1, x_2 \in \mathcal{C}_q, \ x_1 \neq x_2 \}$.
For each pair $(x_1, x_2)$, we prompt the LLM to compare $x_1$ and $x_2$ and provide justification for why one is preferable over the other. To ensure consistency, we repeat this process three times and select the majority decision as the final output. This process results in 900,000 pairwise comparisons, each accompanied by a justification. The prompt used is illustrated in Figure~\ref{fig:pairwise-prompt}.

To convert the pairwise comparisons into a listwise ranking, we utilize the Bradley-Terry model~\cite{19ff28b9-64f9-3656-ba40-08326a05748e} and the Plackett-Luce model~\cite{af5079a1-8ca5-3727-a405-0a82390327b7}. To evaluate the correlation between plausibility scores from the \textit{Candidate Generation} and \textit{Pairwise Comparison} stages, we compute Spearman correlation ~\cite{ca468a70-0be4-389a-b0b9-5dd1ff52b33f}, Pearson correlation~\cite{1320776d-9e76-337e-a755-73010b6e4b64} and KL-Divergence~\cite{a0dc553c-0830-3fe2-ab2a-eece0d66a7db}. The results of these correlation analyses are shown in Table~\ref{tbl:listwise-pairwie-correlation}.
They indicate a strong positive correlation between listwise-based plausibility scores and pairwise-based plausibility scores across both models, as measured by Spearman correlation, Pearson correlation and KL-Divergence.

\begin{figure}
	\centering
	\includegraphics[width=0.9\columnwidth]{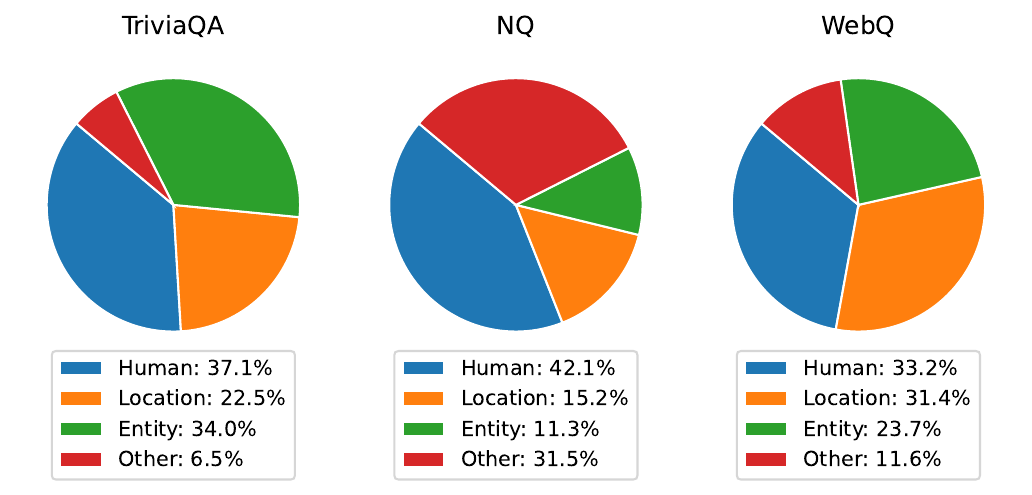}
	\caption{The distribution of the \datasetname dataset across the TriviaQA, NQ, and WebQ datasets.}
	\label{fig:distribution}
        \Description{}
\end{figure}

\begin{table}
	\caption{KL-Divergence, Spearman, and Pearson correlation coefficients between plausibility scores from the Candidate Generation and Pairwise Comparison stages, based on the Bradley-Terry and Plackett-Luce models.}
	\label{tbl:listwise-pairwie-correlation}
	\centering
	\resizebox{\columnwidth}{!}{%
		\scriptsize
		\begin{tabular}{@{}lccc@{}}
			\toprule
			Model                                                                 & KL-Divergence & Spearman      & Pearson       \\ \midrule
			Bradley-Terry~\cite{19ff28b9-64f9-3656-ba40-08326a05748e}             & 1.34          & 0.62          & 0.62          \\
			Plackett-Luce~\cite{af5079a1-8ca5-3727-a405-0a82390327b7}             & 1.17          & 0.62          & 0.64          \\
			Initialized-Plackett-Luce~\cite{af5079a1-8ca5-3727-a405-0a82390327b7} & \textbf{1.16} & \textbf{0.65} & \textbf{0.65} \\ \bottomrule
		\end{tabular}%
	}
\end{table}

In the \datasetname dataset, we include both listwise-based and pairwise-based plausibility scores, allowing users to analyze both of them or to choose one preferred type of scores. Each strategy has its own advantages and drawbacks. Listwise-based plausibility scores require fewer computations and resources since they are produced using a single prompt, whereas pairwise-based plausibility scores require significantly more computations and resources, growing quadratically compared to the first strategy. 
From another perspective, listwise-based plausibility scores tend to be somewhat artificial. When manually assessed, listwise-based plausibility scores often appear in rounded multiples of 5, such as 10, 15, or 25, making them appear less precise. In contrast, pairwise-based plausibility scores are more fine-grained and realistic, frequently including decimal values that better reflect subtle distinctions in plausibility.

\subsection{Preparation}\label{ss:preparation}
This module gathers the questions, candidate answers, listwise-based plausibility scores, and pairwise-based plausibility scores, enriching the dataset with additional attributes to create the final version.

\subsubsection{Pairwise To Listwise}\label{sss:pairtolist}
Based on the results in Table~\ref{tbl:listwise-pairwie-correlation}, which demonstrate a high correlation between listwise-based and pairwise-based plausibility scores, we conclude that listwise-based plausibility scores can be used as initialized values for the Plackett-Luce model~\cite{af5079a1-8ca5-3727-a405-0a82390327b7} for producing the final pairwise-based plausibility scores that we refer as Initialized Plackett-Luce. This approach leads to more accurate and highly correlated plausibility scores for candidate answers. The results of the Initialized Plackett-Luce model are presented in Table~\ref{tbl:listwise-pairwie-correlation}, showing that the initialized version achieves a higher correlation than other models.

\begin{table}
	\caption{The attributes of the \datasetname dataset. In pairwise comparisons, $x_1$ and $x_2$ represent the first and second candidate answers, respectively.}
	\label{tbl:dataset_attributes}
	\resizebox{\columnwidth}{!}{%
		\begin{tabular}{@{}lll@{}}
			\toprule
			\textbf{Attribute}                  & \textbf{Sub-Attribute} & \textbf{Description}                                             \\ \midrule
			id                                  & -                      & Unique identifier for the question                               \\ \midrule
			question                            & -                      & The text of the question                                         \\ \midrule
			\multirow{2}{*}{question\_type}     & major                  & Coarse-grained class of the question type                        \\
			                                    & minor                  & Fine-grained class of the question type                          \\ \midrule
			question\_difficulty                & -                      & A measure of how difficult the question is to answer             \\ \midrule
			answer                              & -                      & The correct answer of the question                               \\ \midrule
			answer\_difficulty                  & -                      & A measure of how difficult the answer is                         \\ \midrule
			\multirow{5}{*}{candidate\_answers} & justification          & Justification for selecting the candidate answer                 \\
			                                    & listwise               & Listwise-based plausibility score                                \\
			                                    & bradley\_terry         & Plausibility score based on the Bradley-Terry model              \\
			                                    & plackett\_luce         & Plausibility score based on the Plackett-Luce model              \\
			                                    & init\_plackett\_luce   & Plausibility score using the Initialized Plackett-Luce model     \\ \midrule
			\multirow{2}{*}{pairwise}           & justification          & Explanation of why $x_1$ is preferred over $x_2$ (or vice versa) \\
			                                    & label                  & Indicates whether $x_1$ is better than $x_2$                     \\ \bottomrule
		\end{tabular}%
	}
\end{table}

\subsubsection{Difficulty Estimation}\label{sss:difficulty_estimation}
For each question, we calculate both question difficulty, which indicates how challenging the question is to answer, and answer difficulty, which reflects how common and well-known the correct answer is.

To assess question difficulty, we adopt the Reference-based Question Complexity approach~\cite{Gabburo2024}, which evaluates difficulty by analyzing how frequently the correct answer appears in retrieved passages and measuring the relevance of these passages to the question. Using these factors, a difficulty score is computed for each question.
For retrieval, we employ the DPR method~\cite{karpukhin-etal-2020-dense}, leveraging the English Wikipedia dump preprocessed by Karpukhin et al.~\cite{karpukhin-etal-2020-dense} as the evidence source. We consider the top 30 most relevant passages to determine the question’s complexity\footnote{Questions with scores below 0.33 are categorized as easy, those above 0.66 as hard, and the remaining as medium.}.

To evaluate answer difficulty, we use the Familiarity evaluation metric~\cite{10.1145/3626772.3657855}, which measures the popularity of entities mentioned in the text by analyzing the number of views on their corresponding Wikipedia pages. This helps determine how well-known the referenced people, places, or concepts are. 
We utilize the spaCy library\footnote{\url{https://spacy.io/}} to identify the entity type of each answer. Next, we employ the Pageview API\footnote{\url{https://github.com/Commonists/pageview-api}} to retrieve the number of views for the Wikipedia pages associated with the extracted entities, covering the period from January 1, 2015, to December 31, 2024. We compute the monthly view counts and normalize them to a range between zero and one to account for high variability in page views.

\subsubsection{Dataset Attributes}\label{sss:dataset_attributes}
Finally, we compile all sampled questions, generated candidate answers with their corresponding justifications, and both listwise-based and pairwise-based plausibility scores along with their justifications. These elements are stored in a JSON file as the final version of \datasetname. Table~\ref{tbl:dataset_attributes} presents the attributes of \datasetname along with their descriptions.

\section{Dataset Analysis}\label{s:dataset_analysis}
In this section, we analyze the \datasetname dataset from two perspectives: \textit{Dataset Statistics} and \textit{Human Evaluation}, which we describe below.

\subsection{Dataset Statistics}\label{ss:dataset_statistics}
Using the pipeline shown in Figure~\ref{fig:pipeline}, we construct a dataset from three popular datasets: TriviaQA \cite{joshi-etal-2017-triviaqa}, NQ~\cite{kwiatkowski-etal-2019-natural}, and WebQ~\cite{berant-etal-2013-semantic}. The dataset comprises 10,000 questions and 100,000 candidate answers, ensuring that each question has 10 candidate answers. Additionally, it includes a total of 1,000,000 justifications. Table~\ref{tbl:dataset_statistics} provides a detailed statistical summary of the dataset across various features.

\begin{table}
	\small
	\caption{Statistical Summary of the \datasetname Dataset}
	\label{tbl:dataset_statistics}
	\resizebox{\columnwidth}{!}{%
		\scriptsize
		\begin{tabular}{@{}l@{\hspace{90pt}}c@{}}
			\toprule
			Metric                                    & Value     \\ \midrule
			Total Questions                           & 10,000    \\
			Total Candidate Answers                   & 100,000   \\
			Total Justifications                      & 1,000,000 \\ \midrule
			Avg. Question Length (words)              & 13.63     \\
			Avg. Answer Length (words)                & 1.35      \\
			Avg. Justification Length (words)         & 274.07    \\ \midrule
			Avg. Number of Entities per Question      & 1.46      \\
			Avg. Number of Entities per Justification & 19.46     \\ \midrule
			Avg. Listwise-based Plausibility Scores   & 31.05     \\
			Avg. Pairwise-based Plausibility Scores   & 48.21     \\ \bottomrule
		\end{tabular}%
	}
\end{table}

\begin{table}
	\centering
	
	\caption{Demographic information of evaluators}
	\resizebox{\columnwidth}{!}{%
		\scriptsize
		\begin{tabular}{@{}c@{\hspace{35pt}}c@{\hspace{35pt}}c@{\hspace{35pt}}c@{}}
			\toprule
			\textbf{Evaluator} & \textbf{Gender} & \textbf{Age} & \textbf{Education Level} \\ \midrule
			1                  & Female          & 18           & High School              \\
			2                  & Female          & 36           & High School              \\
			3                  & Female          & 30           & Bachelor's Degree        \\
			4                  & Male            & 45           & Bachelor's Degree        \\
			5                  & Male            & 40           & Master's Degree          \\
			6                  & Male            & 33           & PhD                      \\ \bottomrule
		\end{tabular}%
	}
	\label{tbl:evaluator_background}
\end{table}

Beyond these statistics, we analyze the distribution of questions based on their source datasets. Figure~\ref{fig:distribution} illustrates the dataset distribution across the three sources and further categorizes the questions by type. The figure shows that most questions focus on the HUMAN category, followed by LOCATION, ENTITY, and OTHER.

\begin{figure}
	\centering
	\includegraphics[width=0.8\columnwidth]{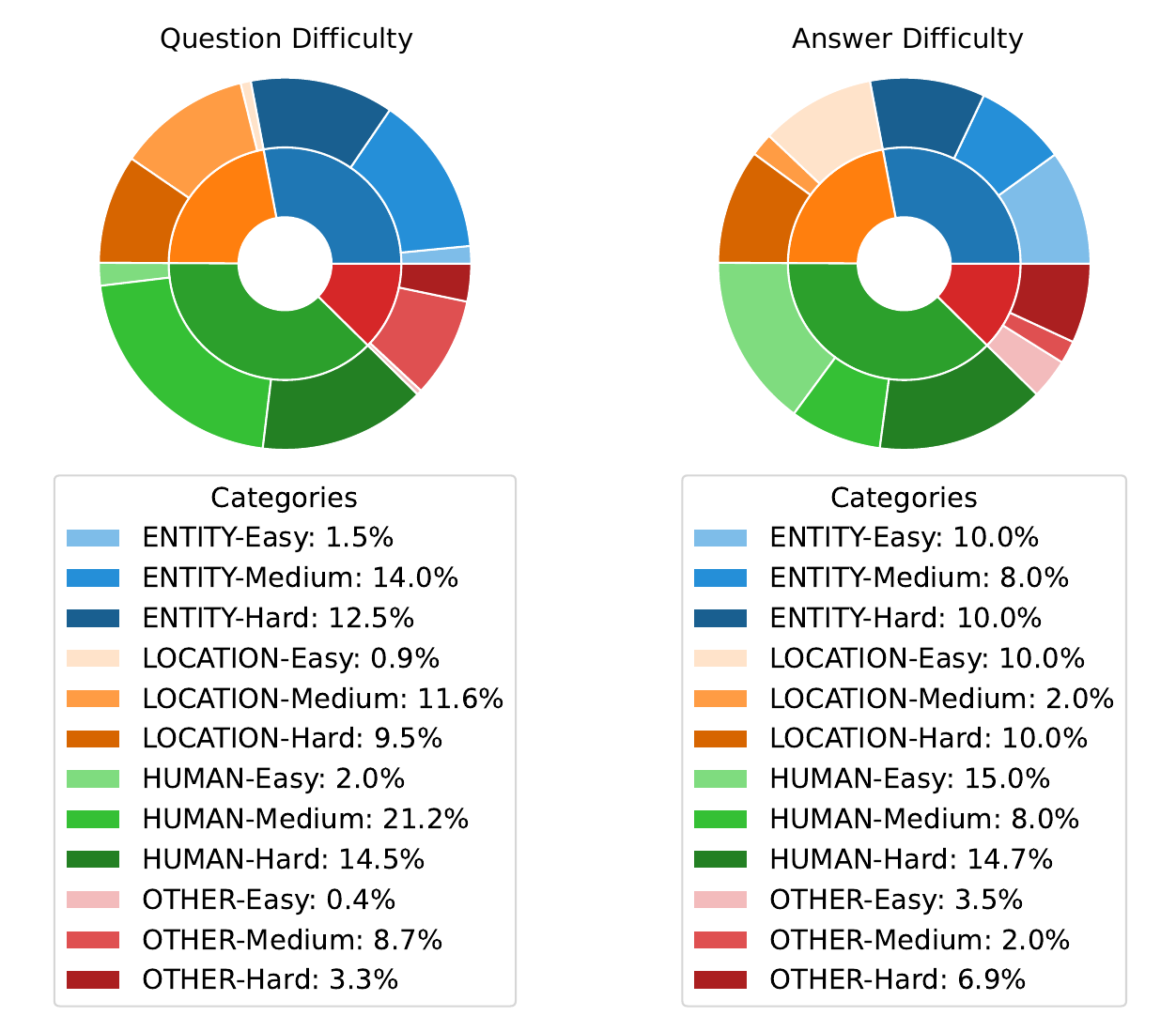}
	\caption{The question and answer difficulty levels in the \datasetname dataset.}
	\label{fig:difficulty}
        \Description{}
\end{figure}

Furthermore, we examine the \datasetname dataset based on question and answer difficulty. Figure~\ref{fig:difficulty} presents the distribution of the dataset across various question types, highlighting both question and answer difficulty levels.

\subsection{Human Evaluation}\label{ss:human_evaluation}
We engaged human evaluators to assess the \datasetname dataset. Our goal was to determine the correlation between the accuracy of the automatic pairwise comparison used in our pipeline and human evaluation results. 

\begin{figure}
	\centering
	\includegraphics[width=\columnwidth]{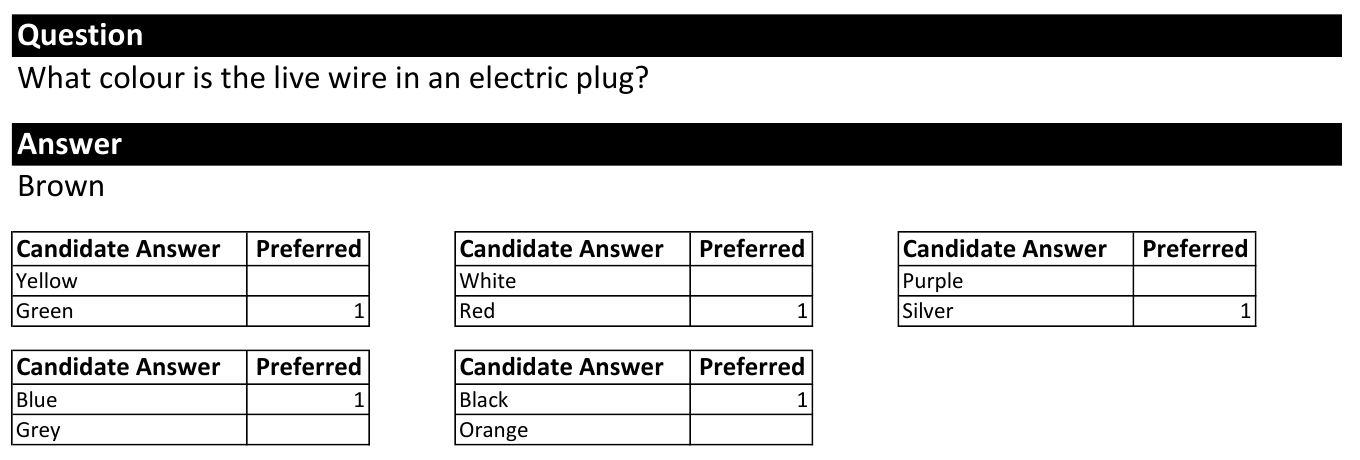}
	\caption{An example of a question annotated by a human evaluator.}
	\label{fig:annotated_example}
        \Description{}
\end{figure}

To achieve this, we selected 100 questions from the dataset using stratified sampling \cite{ARNAB2017213}, ensuring that the distribution of question types and answer difficulty remained consistent. For each question, we generated five unique pairs from the candidate answers. The questions along with their pairs were then divided into two sets of 50. Each set was evaluated by three different individuals, who were asked to compare the answer pairs and determine which option was more plausible as the correct answer while still being an incorrect choice.
It is important to note that within each set, the same questions and corresponding answer pairs were presented to all three evaluators. In other words, each group of three individuals was asked to evaluate the same set of questions and answer pairs. Figure~\ref{fig:annotated_example} presents an example of a question evaluated by human annotators.

\begin{figure*}
	\centering
	\includegraphics[width=0.8\textwidth]{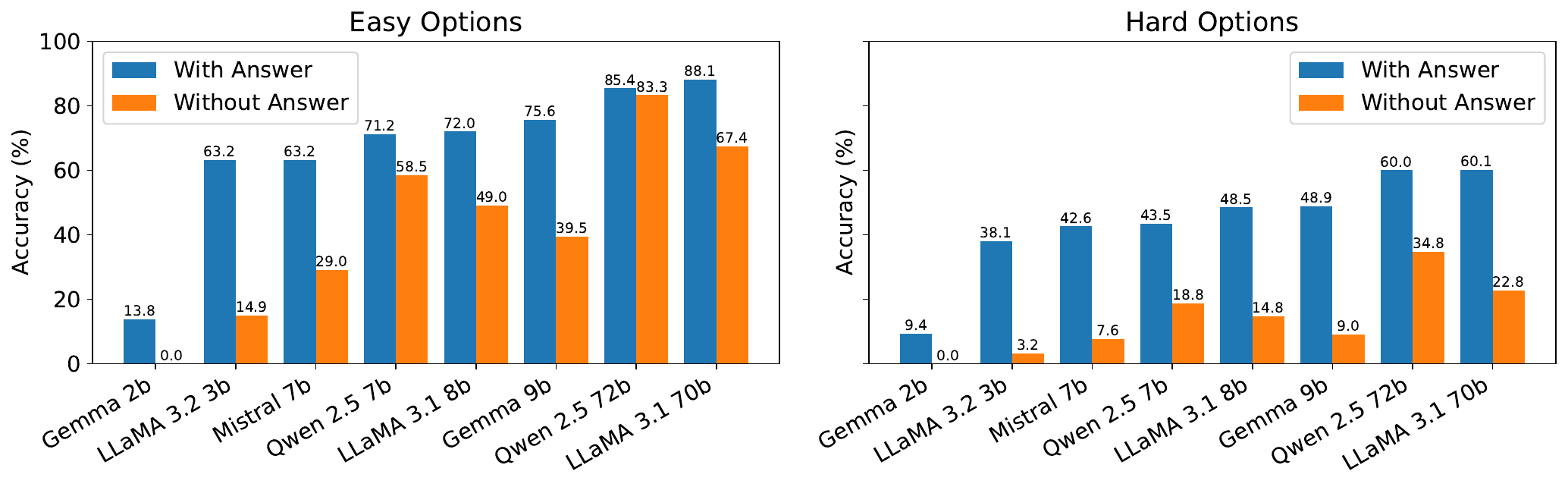}
	\caption{Accuracy of MCQA systems when the options are either easy or hard, and when the correct answer is included among the options or replaced with \textit{None of them}.}
	\label{fig:mcqa_results}
        \Description{}
\end{figure*}

\begin{figure}[t]
	\centering
	
	\begin{center}
		\fcolorbox{customblue}{framegray}{
			\begin{minipage}{0.88\linewidth}
				\scriptsize
				You are a knowledgeable assistant. Answer the following multiple-choice question by selecting the most appropriate option.
						
				Question: $<$question$>$
				\begin{enumerate}[label=\Alph*)]
					\item $<$option\_1$>$
					\item $<$option\_2$>$
					\item $<$option\_3$>$
					\item $<$option\_4$>$
				\end{enumerate}
				        		    
				The letter of the chosen option must be from [A, B, C, D]. Format your response as a JSON item, where each output is represented as:
								
				\begin{flushleft}
					\ttfamily
					\{\\
					\ \ "Letter": "<letter>"\\
					\ \ "Option": "<option>"\\
					\}
				\end{flushleft}
				\vspace{\baselineskip}
				The output must be only a JSON item.
			\end{minipage}
		}
	\end{center}
	\caption{Prompt for MCQA. \texttt{<question>} represents the given question, while each of the four candidate answers is denoted by \texttt{<option\_1>} to \texttt{<option\_4>}. \texttt{<letter>} and \texttt{<option>} indicate the letter and corresponding value that the LLM identifies as the correct answer.}
	\label{fig:mcqa-prompt}
	\Description{}
\end{figure}

The instructions given to the evaluators were as follows:

\begin{enumerate} \item Start by thoroughly reviewing the question, correct answer, and candidate answer pairs in the Excel sheets.
	\item Assess whether you are familiar with the correct answer and the candidate answers.
	\item If familiar, compare the answers in each pair and select the more plausible one, annotating your choice in the sheet.
	\item If unfamiliar, research the content online to understand the correct and candidate answers before evaluating.
	\item Use designated columns in the Excel sheet for your selections.
\end{enumerate}

Table~\ref{tbl:evaluator_background} provides an overview of the human evaluators' backgrounds, including their academic levels, genders, and age groups, ensuring a diverse and representative participant pool.  

To analyze the human assessment results, we consider the majority selection for each answer pair as the final human annotation. These annotations are then compared with the pairwise labels in the dataset to compute accuracy. Table~\ref{tbl:human_results} presents the human evaluation results across different difficulty levels. It illustrates the accuracy of our pipeline in identifying the better candidate answer in a pairwise comparison with human annotations considered as the ground truth. The results for \textit{Entire} indicate that our pipeline effectively identifies better candidate answers, demonstrating high agreement with human judgments. Furthermore, the table highlights that selecting the better candidate answer becomes more challenging as the difficulty level increases, with easier answers being more distinguishable than harder ones.

\section{Experiments and Results}\label{s:experiments}
In this section, we describe the experiments conducted on two different tasks—MCQA and QARA—to demonstrate the applicability of the \datasetname dataset across various IR/NLP subtasks. Due to the time-consuming character of the experiments, we sample 1,000 questions from the dataset as the test set using the stratified sampling method \cite{ARNAB2017213}, ensuring a balanced distribution based on question types, question difficulty, and answer difficulty. In the experiments, we use the \textit{Initialized-Plackett-Luce} scores as plausibility scores.

\subsection{MCQA Results}\label{ss:mcqa_results}
One of the primary applications of candidate answers with plausibility scores is in the MCQA task, particularly for distractor generation. This task involves generating high-quality and challenging distractors for MCQA systems. Depending on the requirement, distractor generation systems produce various distractors with different difficulty levels.  

We use the candidate answers in the \datasetname dataset as distractors for MCQA systems and treat LLMs as the MCQA models. In this experiment, we set the number of answer options to four. To assess the quality of the candidate answers, we conduct four different experiments as follows:

\begin{figure}[t]
	\centering
	
	\begin{center}
		\fcolorbox{customblue}{framegray}{
			\begin{minipage}{0.88\linewidth}
				\scriptsize
				Question: <question>
						    
				Answer: <can>
				\newline
				        		    
				Is answer correct? Select either "Yes" or "No".
			\end{minipage}
		}
	\end{center}
	    
	\caption{Prompt for verifying whether a candidate answer is correct. \texttt{<question>} represents the given question, and \texttt{<can>} denotes the candidate answer being evaluated.}
	\label{fig:robustness-prompt}
	\Description{}
\end{figure}

\begin{table}
	\caption{Human Evaluation Results for the \datasetname Dataset. The table shows the accuracy of our pipeline in detecting the better candidate answer in pairwise comparisons when using human evaluation results as ground truth.}
	\label{tbl:human_results}
	\resizebox{\columnwidth}{!}{%
		\scriptsize
		\begin{tabular}{@{}l@{\hspace{80pt}}cccc@{}}
			\toprule
			Difficulty & Entire & Easy    & Medium  & Hard    \\ \midrule
			Accuracy   & 79.2\% & 87.06\% & 81.21\% & 69.09\% \\ \bottomrule
		\end{tabular}%
	}
\end{table}

\begin{enumerate}  
	\item \textbf{Hard-wA}: The options are difficult, meaning they are the top three highly-plausible candidate answers, and the correct answer is included as the remaining option.  
	\item \textbf{Hard-woA}: The options are difficult, meaning they are the top three highly-plausible candidate answers, and the  remaining option is \textit{None of them}.  
	\item \textbf{Easy-wA}: The options are easy, meaning they are the bottom three plausible candidate answers, and the correct answer is included as the remaining option.  
	\item \textbf{Easy-woA}: The options are easy, meaning they are the bottom three plausible candidate answers, and the  remaining option is \textit{None of them}.
\end{enumerate}  

The purpose of using difficult (\textit{Hard-wA} and \textit{Hard-woA}) and easy (\textit{Easy-wA} and \textit{Easy-woA}) options is to assess the accuracy of plausibility scores and their ability to challenge MCQA systems in identifying the correct answer from the options. Additionally, comparing conditions where the correct answer is present in the options (\textit{Hard-wA} and \textit{Easy-wA}) versus cases where \textit{None of them} is included (\textit{Hard-woA} and \textit{Easy-woA}) helps evaluate how the presence of the correct answer impacts MCQA system performance.  

We use eight different LLMs from various model families and sizes to assess the effectiveness of the \datasetname dataset across different scenarios including Gemma-2b~\cite{2024arXiv240800118G}, LLaMA-3.2-3b~\cite{2024arXiv240721783G}, Mistral-7b~\cite{2023arXiv231006825J}, Qwen-2.5-7b~\cite{2024arXiv241215115Q}, LLaMA-3.1-8b~\cite{2023arXiv230213971T}, Gemma-9b~\cite{2024arXiv240800118G}, LLaMA-3.1-70b~\cite{2023arXiv230213971T}, and Qwen-2.5-72b~\cite{2024arXiv241215115Q}. Figure~\ref{fig:mcqa-prompt} illustrates the prompt used in LLMs for the MCQA system.

\begin{figure}[t]
	\centering
	\includegraphics[width=0.7\columnwidth]{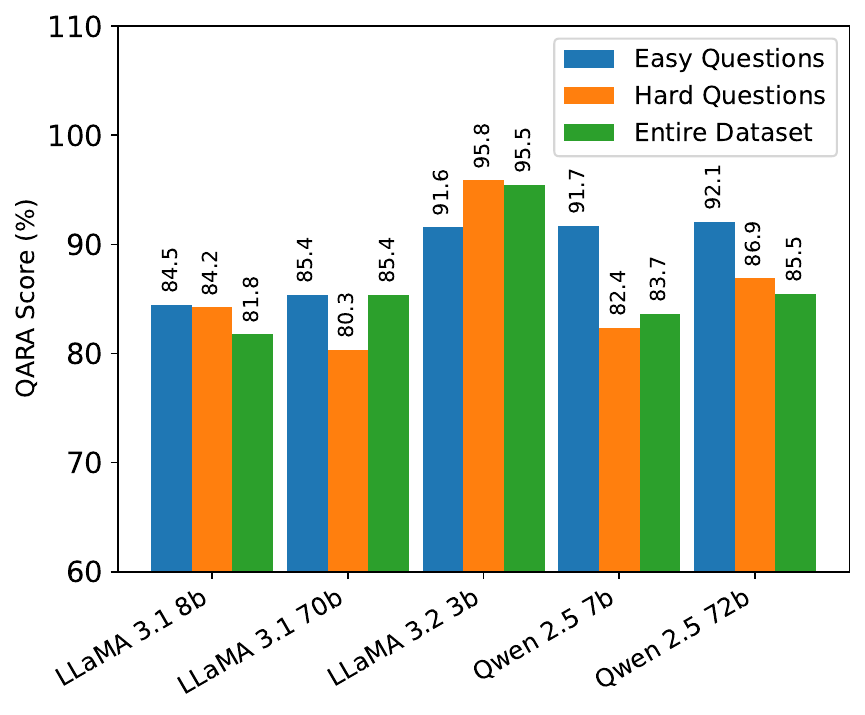}
	\caption{The results of QARA for five different LLMs across the entire dataset, easy questions, and hard questions.}
	\label{fig:robustness_results}
        \Description{}
\end{figure}

Figure~\ref{fig:mcqa_results} demonstrates that identifying the correct answer is more challenging when the difficulty level of the options is high compared to when the options are easy. This highlights the reliability and realism of the plausibility scores assigned to candidate answers. Additionally, the figure shows that the presence of the correct answer among the options improves the accuracy of the MCQA system compared to when it is absent and \textit{None of them} is the correct choice. The results also indicate that the MCQA system tends to select one of the candidate answers as correct, rather than choosing \textit{None of them}, further demonstrating the quality of the candidate answers and their plausibility for the question, even in cases where they are ultimately incorrect.

\subsection{QA
	Robustness Results}\label{ss:robustnessqa_results}
Candidate answers with plausibility scores can also be utilized in QA
Robustness Assessment (QARA) of models. This task assesses how well a QA system can remain robust against incorrect answers when asked to determine their correctness.  

In this experiment, we treat the candidate answers from the \datasetname dataset as incorrect answers and evaluate the QARA score. LLMs are used as QA systems for this experiment. First, we prompt an LLM to verify whether the candidate answers for each question are correct. Figure~\ref{fig:robustness-prompt} presents the prompt we use. This prompt generates either \textit{Yes} or \textit{No} for each question-candidate answer pair. Based on the LLM-generated responses, we define the function $\chi_q(c)$, which indicates whether the QA system is robust against candidate answer $c$ for the question $q$:

\begin{equation}
	\chi_q(c) = \left\{ \begin{array}{ll}
	0 & if~QA(q,c)=Yes \\
	1 & if~QA(q,c)=No
	\end{array} \right.
\end{equation}

Then, using the $\chi_q(c)$ function, we calculate the robustness score for the question $q$:

\begin{equation}
	\mathcal{R}(q) = \frac{\sum_{c \in \mathcal{C}_q}{\mathcal{P}(c).\chi_q(c)}}{\sum_{c \in \mathcal{C}_q}{\mathcal{P}(c)}}
\end{equation}

\begin{table}
	\caption{Spearman and Pearson correlation coefficients between QARA and ExactMatch, as well as QARA and Contains, for the \datasetname dataset.}
	\label{tbl:robustness_correlation}
	\resizebox{\columnwidth}{!}{%
		\scriptsize
		\begin{tabular}{@{}ll@{\hspace{70pt}}c@{}}
			\toprule
			Method                    & Metric     & \multicolumn{1}{l}{Correlation Coefficient} \\ \midrule
			\multirow{2}{*}{Spearman} & ExactMatch & -0.1                                        \\
			                          & Contains   & 0.1                                         \\ \midrule
			\multirow{2}{*}{Pearson}  & ExactMatch & -0.07                                       \\
			                          & Contains   & -0.28                                       \\ \bottomrule
		\end{tabular}%
	}
\end{table}

\begin{table}
	\caption{The success rate of various LLMs in demonstrating robustness against candidate answers.}
	\label{tbl:robustness_successrate}
	\resizebox{\columnwidth}{!}{%
		\scriptsize
		\begin{tabular}{@{}l@{\hspace{20pt}}c@{\hspace{70pt}}c@{}}
			\toprule
			Model     & \# of Parameters & SuccessRate (\%) \\ \midrule
			LLaMA 3.1 & 8b               & 30.2             \\
			Qwen 2.5  & 7b               & 33.06            \\
			LLaMA 3.1 & 70b              & 34.29            \\
			Qwen 2.5  & 72b              & 36.38            \\
			LLaMA 3.2 & 3b               & 46.2             \\ \bottomrule
		\end{tabular}%
	}
\end{table}

Where $\mathcal{C}_q$ denotes the candidate answers for question $q$, and $\mathcal{P}(c)$ represents the plausibility score of the candidate answer $c$.  
In simpler terms, the robustness score of a QA system for a question is calculated as the sum of the plausibility scores of the candidate answers that the QA system correctly identifies as wrong, divided by the total plausibility scores of all candidate answers for the question.  
Finally, we compute the overall QARA score for the QA system using:

\begin{equation}
	QARA = \frac{\sum_{q \in Q_{c}}{\mathcal{R}(q)}}{\lvert Q_{c}\rvert}
\end{equation}

where $Q_{c} = \{q \in Q \mid QA(q) \in A_q\}$, indicating that we only consider the questions for which the QA system's answer belongs to the set of correct answers $A_q$. We do this because including questions that QA system cannot answer would be unfair when evaluating its robustness. In other words, this approach evaluates how robust a QA system is against candidate answers for questions it already knows the correct answer to.

In this experiment, we use five LLMs as QA systems, including LLaMA-3.2-3b, Qwen-2.5-7b, LLaMA-3.1-8b, LLaMA-3.1-70b, and Qwen-2.5-72b. We chose these models because they represent a range of sizes and come from different LLM families\footnote{We did not include Gemma and Mistral because this experiment is time-consuming, and we needed to exclude some models.}.

To examine whether QARA can be computed using ExactMatch and Contains metrics, we calculate the Spearman and Pearson correlation coefficients between QARA and ExactMatch and Contains. Table~\ref{tbl:robustness_correlation} presents the results. The findings show no significant correlation between QARA and ExactMatch or Contains, indicating that QARA is an independent factor for evaluating the robustness of QA systems.

\begin{figure}
	\centering
	\includegraphics[width=0.7\columnwidth]{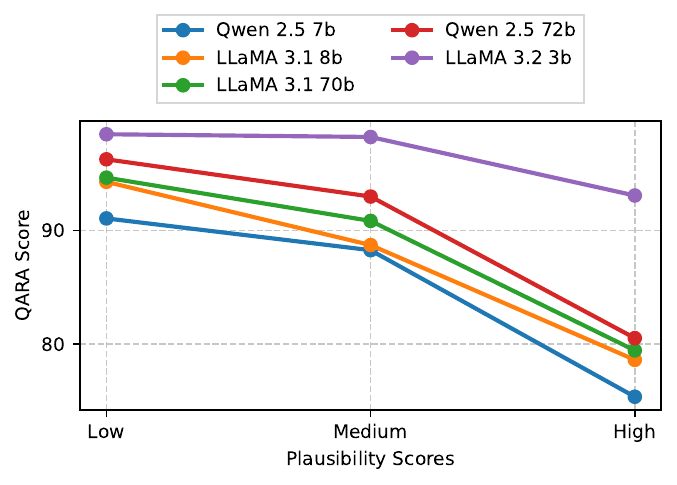}
	\caption{The results of QARA of LLMs when using candidate answers with three different levels of plausibility scores including \texttt{Low} (0-33), \texttt{Medium} (34-66), and \texttt{High} (67-100). }
	\label{fig:plausible_qara}
        \Description{}
\end{figure}

Additionally, Figure~\ref{fig:robustness_results} shows the QARA
results for the entire dataset, easy questions, and hard questions across five LLMs as QA systems. The results indicate that QA systems are less robust for hard questions than for easy questions. This may be because LLMs are less confident when answering hard questions, whereas, for easy questions, they tend to be more certain of their answers and can more easily identify incorrect candidate answers.

Furthermore, we evaluate the success rate of each QA system, which measures the percentage of questions for which the system knows the answer and for which it is robust against incorrect candidate answers. Table~\ref{tbl:robustness_successrate} presents the success rates for various LLMs as QA systems. 
This evaluation underscores the differing capabilities of LLMs in managing complex question-answering tasks, which may be influenced by their training strategies. The variations in success rates among these systems not only suggest a potential impact of how each model has been trained to handle and reject incorrect candidate answers but also reflect the effectiveness of their underlying algorithms in discerning and filtering out less accurate responses. 

Finally, we analyze the QARA of various LLMs across different levels of answer plausibility. These include \texttt{Low}, comprising candidate answers with plausibility scores ranging from 0 to 33, \texttt{Medium}, featuring scores from 33 to 66, and \texttt{High}, containing answers with scores from 66 to 100. Figure~\ref{fig:plausible_qara} displays the performance of different LLMs against these categories.
The results confirm that the plausibility scores of candidate answers in \datasetname dataset are effective for QA robustness analysis. They demonstrate that LLMs exhibit greater robustness against candidate answers with decreasing levels of plausibility. The answers with higher plausibility scores prove more challenging for LLMs to be identified as incorrect compared to those with lower scores. This finding suggests the effectiveness of our evaluation pipeline in terms of accurately determining the plausibility degrees of candidate answers.

\section{Limitations and Use Cases}\label{s:use_cases}

While our approach introduces a novel dataset and pipeline for generating candidate answers with plausibility scores, it has certain limitations:

\begin{itemize}
	\item The dataset focuses exclusively on factoid questions, limiting its applicability to open-ended or reasoning-based queries.
	\item Since the pipeline relies on LLMs, any biases or limitations inherent in these models may be reflected in the output.
	\item The experiments conducted are based on a fixed set of 10 candidate answers per question, which may not capture variations in answer plausibility beyond this scope.
\end{itemize}

Despite these limitations, \datasetname and the proposed pipeline support various applications in QA evaluation, natural language processing, and information retrieval:

\begin{itemize}
	\item It could enable the generation of realistic and challenging multiple-choice options at varying difficulty levels, improving MCQA distractor generation.
	\item It could enhance QARA by measuring model resilience in rejecting highly plausible yet incorrect answers.
	\item Plausibility scores could improve negative example selection in contrastive learning, contributing to better model training~\cite{cao-etal-2022-exploring}.
	\item The entropy of plausibility scores could aid question difficulty estimation, facilitating dynamic complexity assessment~\cite{10.1145/3556538}.
	\item The dataset could support research in automatic hint generation and evaluation, particularly for assessing the Convergence metric~\cite{mozafari2025hinteval}, which measures how effectively hints narrow down potential answers to a question. The resulting scores could contribute to a more accurate assessment of convergence within the HintEval framework\footnote{\url{https://github.com/DataScienceUIBK/HintEval}}.
\end{itemize}

\section{Conclusion}\label{s:conclusion}
In this paper, we introduced \datasetname, a large-scale QA dataset that incorporates plausibility scores for candidate answers, providing a novel resource for MCQA and QARA. The dataset consists of 10,000 questions and 100,000 candidate answers, each annotated with its plausibility score. We validated the dataset through human evaluations and statistical analyses, demonstrating its reliability and applicability.
Moreover, the proposed pipeline is highly adaptable and can be easily applied to other QA datasets, allowing researchers produce similar resources for a wider range of applications.

We conducted extensive experiments to showcase the usefulness of \datasetname. Our results indicate that plausibility scores effectively enhance the quality of MCQA distractors, making them more challenging for QA models. QA robustness scores reveal that existing large language models (LLMs) struggle to differentiate between highly plausible incorrect answers and correct ones, highlighting the importance of robustness evaluation. 

For future research, we will investigate fine-tuning LLMs with \datasetname to improve their ability to distinguish plausible yet incorrect answers. Additionally, we will investigate the potential of using answer plausibility as a metric for assessing question difficulty.

\bibliographystyle{ACM-Reference-Format}
\balance
\bibliography{Main}

\end{document}